# Approximate spectral clustering density–based similarity for noisy datasets

Mashaan Alshammari ∗ and Masahiro Takatsuka

*School of Computer Science, The University of Sydney, NSW 2006, Australia*

ABSTRACT

Approximate spectral clustering (ASC) was developed to overcome heavy computational demands of spectral clustering (SC). It maintains SC ability in predicting non-convex clusters. Since it involves a preprocessing step, ASC defines new similarity measures to assign weights on graph edges. Connectivity matrix (CONN) is an efficient similarity measure to construct graphs for ASC. It defines the weight between two vertices as the number of points assigned to them during vector quantization training. However, this relationship is undirected, where it is not clear which of the vertices is contributing more to that edge. Also, CONN could be tricked by noisy density between clusters. We defined a directed version of CONN, named DCONN, to get insights on vertices contributions to edges. Also, we provided filtering schemes to ensure CONN edges are highlighting potential clusters. Experiments reveal that the proposed filtering was highly efficient when noise cannot be tolerated by CONN.

## 1. Introduction

Spectral clustering emerged to be an efficient learning paradigm. It broadens the definition of clustering process from “points that share a common mean” to “points that are strongly connected”. This definition enables spectral clustering to identify more complex shaped clusters. However, the gains of spectral clustering are usually outweighed by its computational demands. Easing the computational demands of spectral clustering was essential to unleash its capabilities. Research efforts brought the concept of approximate spectral clustering (ASC). It operates by choosing a subset ($m$ representatives) from the data ($n$ points) and carry on with spectral clustering steps. Eventually, the outcome of $m$ points will be generalized to all $n$ points (Chen and Cai, 2011; Yan et al., 2009).

The most important question about ASC was: how to define similarities between $m$ representatives. Distance-based, density-based, or hybrid similarities were the most used. Distance-based similarities are criticized for including decaying parameters (Tasdemir, 2012). Defining similarities based on density is more accurate than distance-based. It creates edges based on continuous density of points between representatives regardless of the distance between them. However, tracking density for each pair of representatives is computationally expensive (Tasdemir et al., 2015).

Connectivity matrix (*CONN*) introduced by (Tasdemir, 2012) minimized the overhead of density tracking by utilizing information from the approximation process. It tracks which representatives are selected as the best matching units (BMUs) and second-best matching units (second BMUs). Then it creates the edges where the weights are the number of points sharing the same BMU. In the case of clean data, *CONN* is superior and it provides an approximated graph that highlights the potential clusters. The problem arises when there is a faulty density (i.e., group of points spanning the area between true clusters) caused by noisy input. *CONN* is guaranteed to create edges in such cases according to its definition.

Faulty density could be determined from the perspective of the two representatives sharing a *CONN* edge ($p$,$q$). In this work, we introduce a directed version of *CONN* named *DCONN*. Examining this directed graph gives us more insights on whether a pair of representatives $p$ and $q$ are mutually agreeing on the density between them. If there is a large difference between the outgoing *DCONN* edge ($p$,$q$) and the incoming *DCONN* edge ($q$,$p$), a red flag is raised on this edge for further inspection. Another test to discover faulty density performed locally by examining whether *CONN* edge (*p*,*q*) falls within the range of acceptance of a specific representative *p*. The acceptance range was defined as the mean of *CONN* edges to all direct neighbors of *p*.

∗ Corresponding author. Tel.: +61 2 9351 3423; fax: +61 2 9351 3838; e-mail: mals6571@uni.sydney.edu.au

We also presented a measure for the mutual agreements of all pair of edges in *DCONN* called (*DCONN Balance*). When *DCONN Balance* is low it means there are small differences between all outgoing and incoming edges in *DCONN*, which means the graph is well balanced. In this case, removing edges will not help because it could break clusters. If *DCONN Balance* is high most representatives are not mutually agreeing on edges. In such cases, removing edges could boost the performance by maintaining mutually agreed on edges. The experimental design revealed that removing edges from *CONN* graph based on the proposed measures could boost the performance of approximate spectral clustering.

This work is organized as follows: in section 2 we review efforts from the literature representing a backbone of this work; the proposed method is introduced in section 3; the experimental setup and discussion are presented in section 4.

## 2. Related work

Spectral clustering operates based on the connectivity of data points instead of compactness required by spherical methods. The idea was initiated back in 1990s and became an attractive clustering tool ever since, a useful summary of earlier efforts could be found in (Weiss, 1999). Given a graph $G = (V,E)$, these efforts were looking for the minimum graph cut between $B$ and $\bar{B}$, $cut(B,\bar{B})$, such that $B \cup \bar{B} = V$ and $B \cup \bar{B} = \phi$. This method was called minimum cut (MinCut). Unfortunately, MinCut tends to partition small and isolated connected components and miss the significant ones because it increases with the number of edges. The notable effort by (Shi and Malik, 2000) introducing the concept of normalized cut (NCut), which was an enhancement over the minimum cut. The term $cut(B,\bar{B})$ was normalized by the total weights from nodes in $B$ and $\bar{B}$ to all nodes in $V$. The proposed formula highlights the significant cut between well-connected components. The second smallest eigenvector of the graph Laplacian $L_{sym}$ represents the value of $Ncut(B,\bar{B})$, where $L_{sym} = D^{-1/2}(D - A)D^{-1/2}$ with $A$ and $D$ are affinity and degree matrices respectively.

The type of constructed graph $G = (V, E)$, heavily influences the performance of spectral clustering. The conventional selection of the graph types involves: fully connected graph, *k*-nearest neighbor graph, and ε-neighborhood graph (von Luxburg, 2007). The former option tends to be inefficient when *n* is large, since it connects all points according to some similarity measure. In *k*- nearest neighbor graphs, each vertex *p* will be connected to *q* if it belongs to *k*-nearest neighbors of *p* (Marchette, 2004). The outcome of such graphs depends heavily on the selection of *k* which needs manual tuning in real applications.

$$(\mathrm{p,q}) \in \mathrm{E(G)} \Leftrightarrow \mathrm{p} \in \mathrm{knn(q)} \text{ or } \mathrm{q} \in \mathrm{knn(p)} \qquad (1)$$

For ε-neighborhood graphs, one simply connects points with pair-wise distances less than ε.

$$(\mathrm{p,q}) \in \mathrm{E(G)} \Leftrightarrow \mathrm{d(p,q)} < \varepsilon \qquad (2)$$

*k*-nearest neighbor and ε-neighborhood graphs involve parameters (i.e., *k* and ε) that require careful tuning. Since these conventional selections of graphs have their limitations, researchers developed various methods to construct the graph $G = (V, E)$ for spectral clustering.

A graph design proposed by (Correa and Lindstrom, 2012) based on empty region graphs (ERGs). ERGs are graphs that connect neighboring points if the geometric region between them does not contain any other points. They proposed a β-skeleton ERG with a local scaling of similarity. The parameter β controls the diameter of balls centered at neighboring points *p* and *q* and their intersection defines the empty region. Furthermore, a graph constructed based on neighborhood density was introduced by (Inkaya, 2015). It operates based on a neighborhood construction (*NC*) by placing a hypersphere passing through the points *p* and *q*. The density of the *NC* is defined by the number of points in the hypersphere, with *p* and *q* directly connected if it was empty and indirectly connected otherwise. Subsequently, the density adaptive neighborhood (*DAN*) graph was constructed as:

$$(\mathrm{p,q}) \in \mathrm{E(DAN)} \Leftrightarrow \mathrm{p} \in \mathrm{NC(q)} \text{ or } \mathrm{q} \in \mathrm{NC(p)} \qquad (3)$$

All aforementioned graph construction methods were primarily developed for spectral clustering. The emerging field of approximate spectral clustering (ASC) was designed to ease the computational demands of spectral clustering while preserving its superiority. ASC starts by selecting a subset *m* out of *n* points, where $m \ll n$, to perform spectral clustering on, then generalize the outcome to all *n*. With this new field, we could still use the previous graph construction methods. Nevertheless, we could be more efficient if we utilize the approximation step as a guide to construct the graph $G = (V, E)$. This was the motivation behind the development of *CONN* graph by (Tasdemir, 2012). It exploits the concept of induced Delaunay triangulation (Martinetz and Schulten, 1994).

$$(\mathrm{p,q}) \in \mathrm{E(CONN)} \Leftrightarrow x_i \in V_{pq} \ or \ x_i \in V_{qp} \qquad (4)$$

where $V_{pq}$ represents the number of points that *p* is their best-matching-unit and *q* second-best-matching-unit.

*CONN* has a couple of advantages that makes it the most suitable graph construction method for ASC with vector quantization. It does not require any parameters, since it uses the outcome of the vector quantization as is. Another advantage of *CONN* is that most of its computations are embedded in vector quantization training. Once the training finished, *CONN* could be trivially computed. Despite these advantages of *CONN*, its reliance on the density to draw edges could impact its accuracy when faulty density spans the void between clusters. This case is more likely to occur in real data since clusters are not well separated. Examining the directed version of *CONN* (named *DCONN*) would give us more insights about what edges to keep. From there we could derive filtered versions of *CONN* with less but more informative edges.

## 3. Proposed Approach

The proposed method aims to filter *CONN* edges that were created by faulty density or wrong positioning of representatives *m*. It starts by constructing a directed version of *CONN* graph, called *DCONN*. Then, it filters all edges where the difference between a pair of *DCONN* edges exceeds a global threshold. A second round of filtering was performed locally. Representatives could cut *CONN* edges that do not align with their local acceptance threshold. Finally, the proposed method performed spectral clustering using the filtered graphs.

### 3.1. DCONN graph

*DCONN* graph is the directed version of *CONN* graph. In *CONN* an edge connecting $p$ and $q$ is defined as in Equation (4). $V_{pq}$ is the set of points where $p$ is the best matching unit (BMU) and $q$ is the second best matching unit. While, $V_{qp}$ is the reversed relationship with points where $q$ is the BMU and $p$ is the second BMU. For *DCONN* graph, a given *CONN* edge was split into a pair of edges defined as:

$$(\mathrm{p}, \mathrm{q}) \in \mathrm{E}(\mathrm{DCONN}) \Leftrightarrow x_i \in V_{pq} \quad (5)$$

$$(\mathrm{q}, \mathrm{p}) \in \mathrm{E}(\mathrm{DCONN}) \Leftrightarrow x_i \in V_{qp} \quad (6)$$

The setup of *DCONN* enables us to uncover which vertex is contributing more in the undirected *CONN* edge (*p*,*q*). For example, if $V_{pq} = 7$ and $V_{qp} = 0$ the *CONN* edge (*p*,*q*) will be 7. For the vertex $q$ this edge is useless because $p$ is not the second BMU for any point assigned to $q$. Given how *DCONN* was constructed the previous relationship will be highlighted, since $(\mathrm{p}, \mathrm{q}) \in \mathrm{E}(\mathrm{DCONN}) = 7$ and $(\mathrm{q}, \mathrm{p}) \in \mathrm{E}(\mathrm{DCONN}) = 0$.

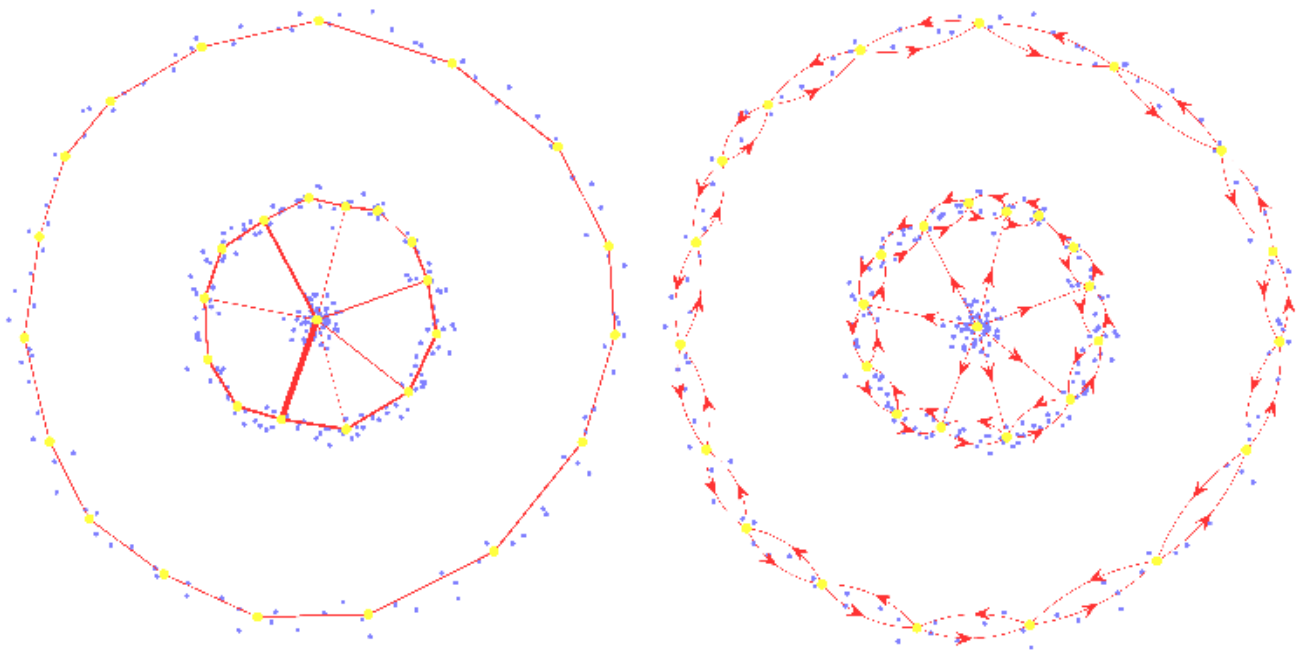

(a) (b)

**Fig. 1. Graphs constructed via *CONN* (a) and *DCONN* (b). All edges from the central representative are wrongly linking two clusters. They were uncovered by *DCONN* definition as unidirectional edges (best viewed in color).**

Fig. 1 shows examples of *CONN* and *DCONN* graphs. In Fig. 1b the representative in the middle was not the second BMU for any representative in the middle ring. This explains why all edges going out of that representative are all unidirectional edges. This is an example of wrong positioning of representatives by the vector quantization method. The one representative in the middle has no neighbors in the same cluster. Therefore, it has to approach representatives in other clusters to satisfy its second BMU condition. Ideally, the *CONN* edges in the middle should not be created since they link two clusters which should be recognized as separate.

### 3.2. Filtering CONN graph

Once *CONN* and *DCONN* graphs are constructed, we are ready to filter the former based on insights from the latter. The first round of filtering removes *CONN* edges with low mutual agreements. If the difference between a pair of *DCONN* edges is higher than a global threshold. This threshold is computed as the mean of all differences between edges pairs in *DCONN* graph plus the standard deviation. For example, if $V_{pq} = 7$ and $V_{qp} = 1$ the difference is 6. If this value is higher than the global threshold ($T_{global}$), the *CONN* edge (*p*,*q*) is removed.

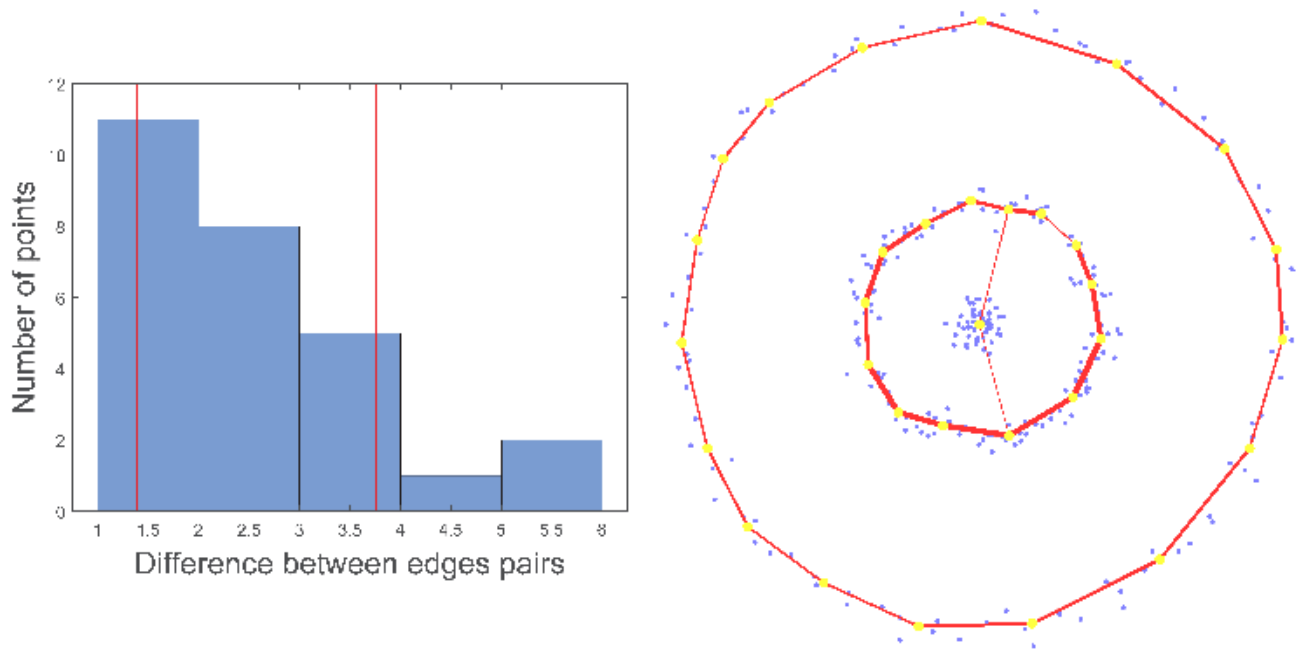

(a) (b)

**Fig. 2. Illustration of first round of filtering (global filtering). (a) histogram of the differences among *CONN* edges pairs, vertical red lines are thresholds (μ ± σ). (b) obtained *CONN*$_G$ graph (best viewed in color).**

An example of a globally filtered graph is shown in Fig. 2. The histogram on the left shows the boundaries of acceptance which is the interval [$\mu \pm \sigma$]. We are concerned about the limit on the right that is slightly below 4. The pair of edges with a difference above that value will be eliminated. This is demonstrated by the graph on the right in which 5 *CONN* edges were eliminated. This globally filtered graph will be referred to as *CONN*$_G$ for the rest of this paper, and its definition is given by equation (7).

The first round of *CONN* filtering was performed globally. This is a deficient setup when the data contains regions with different statistics. Therefore, it was rectified by a second round of filtering performed locally. Given a vertex $p$ in *CONN* graph, it could be connected to multiple *CONN* edges. The local threshold was set as the mean of distances to all vertices connected to $p$ plus the standard deviation. If a *CONN* edge exceeds the local threshold (i.e., $T_{local}$) it will get a vote for removal by $p$. Two graphs were derived using this filtering. The first graph removes all *CONN* edges with 1 removal vote, while the second removes edges with 2 removal votes.

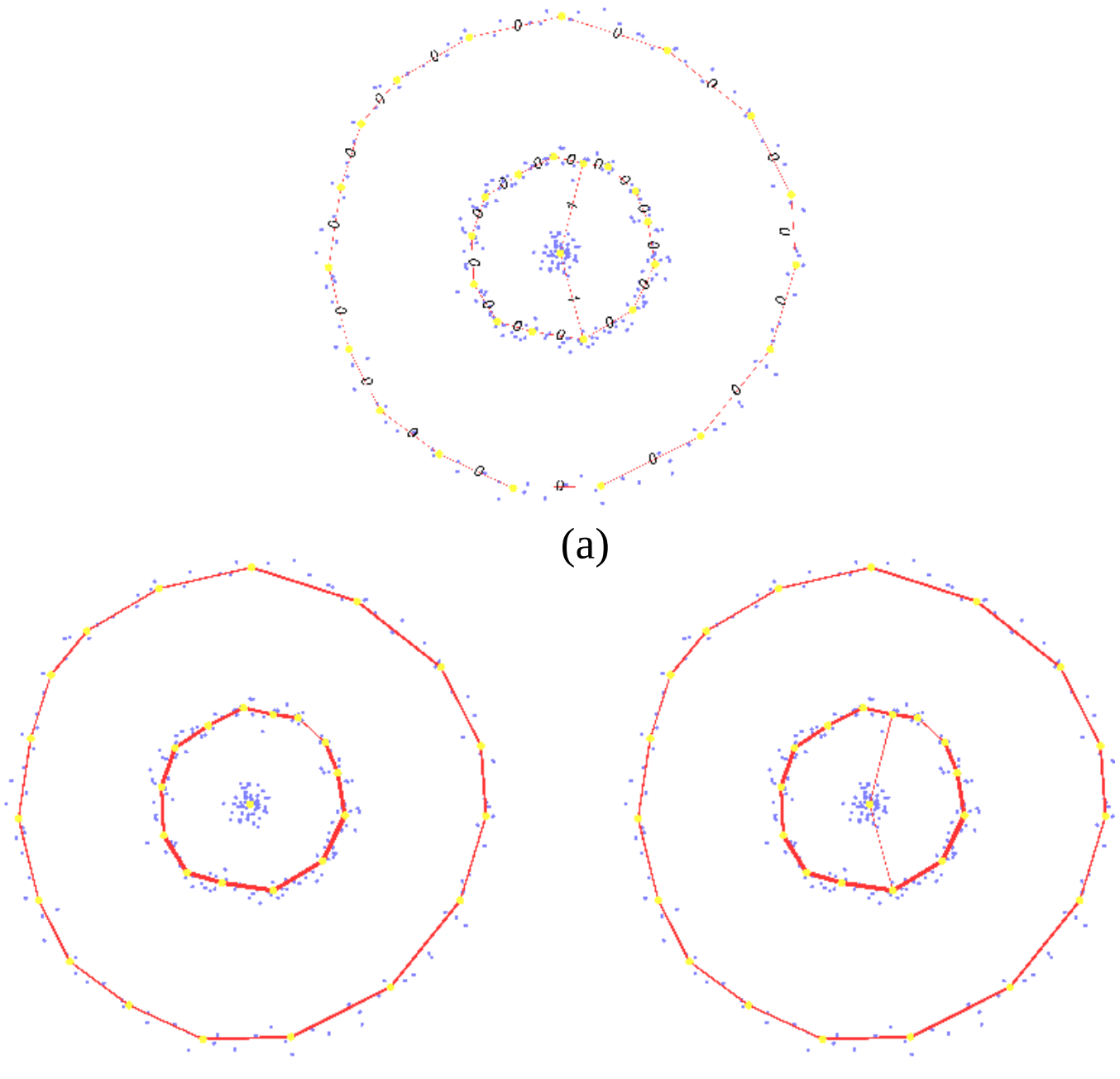

(a)

(b) (c)

**Fig. 3. Illustration of second round of filtering (local filtering). (a) voting scheme based on local threshold (0: edge is within acceptance, 1: edge is violating one vertex threshold, 2: edge is violating both vertices thresholds). (b) *CONN*$_{L1}$ graph removes all edges with 1 vote. (c) *CONN*$_{L2}$ graph removes all edges with 2 votes (best viewed in color).**

$$(p,q) \in E(CONN_G) \Leftrightarrow (x_i \in V_{pq} \text{ or } x_i \in V_{qp}) \text{ and } (|V_{pq} - V_{qp}| < T_{global}) \quad (7)$$

$$(p,q) \in E(CONN_{L1}) \Leftrightarrow (p,q) \in E(CONN_G) \text{ and } (d(p,q) < T_p \text{ and } d(p,q) < T_q) \quad (8)$$

$$(p,q) \in E(CONN_{L2}) \Leftrightarrow (p,q) \in E(CONN_G) \text{ and } (d(p,q) < T_p \text{ or } d(p,q) < T_q) \quad (9)$$

Examples of aforementioned graphs are illustrated in Fig. 3. Fig. 3a is the voting scheme highlighted as weights on edges. In Fig. 3b, all the remaining faulty edges escaped the global filtering were removed, since they received one vote each. In Fig. 3c, these edges were kept since they did not receive two votes. This is understood since the neuron on the middle would not vote for removing these edges because it only has these two edges. Both graphs will be referred to as $CONN_{L1}$ and $CONN_{L2}$ in upcoming sections, and their definitions are given by equations (8) and (9) respectively.

### *3.3. DCONN balance measure*

*DCONN* is an addition on top of *CONN*, hence, it requires additional computations. Anticipating when the proposed method would outperform *CONN* helps to avoid unnecessary computations and contributes the efficiency of this proposal. This section introduces a proposed measure to evaluate *CONN* performance without carrying out all computations described in the previous section.

The ultimate objective of *CONN* filtering proposed previously is removing edges with low mutual agreement. The lowest level of mutual agreement is in form of unidirectional *DCONN* edge. A *CONN* graph with high mutual agreement would have lower differences among its directed edges (i.e., *DCONN*). Computing these differences and normalizing them by weights on edges would give us a hint about the over- all mutual agreements in *CONN* graph. We call this score *DCONN Balance*

Fig. 4 shows two examples of the same dataset with different noise levels. In the clean version, *DCONN* produces a well-balanced graph highlighting potential clusters. The gains obtained through filtering this graph will be outweighed by the additional computations. On the other hand, the noisy graph has multiple unidirectional edges. Removing these edges will ensure better clustering accuracy, therefore, additional computations should not be a barrier.

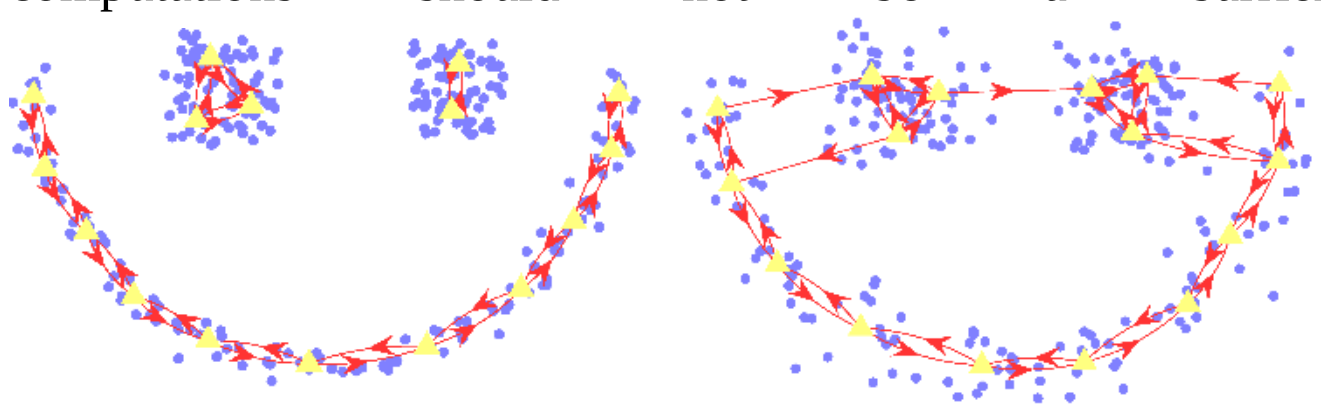

(a) (b)

**Fig. 4. Constructing *DCONN* graph for clean and noisy versions of a synthetic dataset. (a) *DCONN Balance* = 0.2105; (b) *DCONN Balance* = 0.3166 (best viewed in color).**

### *3.4. Performing Spectral Clustering*

Spectral clustering steps performed on the prepared graphs: *CONN*, $CONN_G$, $CONN_{L1}$, and $CONN_{L2}$. It represents each graph by its corresponding affinity matrix where $A(i, j)$ is defined as:

$$A_{pq} = (\mathrm{p,q}) \in \mathrm{E(CONN)}, \quad where\ p \neq q \quad (10)$$

$$\mathrm{L_{sym}} = \mathrm{D^{-1/2}LD^{-1/2}} = \mathrm{I} - \mathrm{D^{-1/2}AD^{-1/2}} \quad (11)$$

The original work of (Ng et al., 2002) used a manually set parameter $k$ to retrieve the top $k$ eigenvectors defining the embedding space $\mathrm{R^{m\times k}}$. We used this approach when the true number of clusters was known (e.g., synthetic data and data from UCI repository). However, when $k$ was unknown like in image segmentation, we used an automatic detection scheme (Alshammari and Takatsuka, 2019). It starts by measuring the separation power of a given eigenvector. The separation was measured using DB index by (Davies and Bouldin, 1979), which is a ratio of intra-cluster and inter-clusters distances. The selected eigenvectors are the ones that have considerably large separation power than other eigenvectors. The embedding space is spanned by the qualifying eigenvectors $\mathrm{R^{m\times k}}$.

## 4. Experiments

The experiments involve different types of examined data: synthetic, images, and data retrieved from UCI repository. For each dataset, four graphs were derived: *CONN*, $CONN_G$, $CONN_{L1}$, and $CONN_{L2}$. Over multiple runs we were mainly concerned about two metrics: clustering accuracy and number of edges in *CONN* graph. The former metric measures the effectiveness of the algorithm given the ground truth labels. The number of edges in *CONN* graph indicates the memory footprint for the algorithm. More edges mean more non-zero elements needed to be stored. We used a windows 10 machine to run the experiments with 3.40 GHz CPU and 8 GB of memory. Algorithms were coded in MATLAB 2017b.

### *4.1. Synthetic data*

Three synthetic datasets were retrieved from the supplementary materials provided by (Zelnik-Manor and Perona, 2005). The objective behind using synthetic data is to test the noise effect on *CONN* graph, moving from a clean to noisy data (see Fig. 5). Before running this experiment, it was anticipated that *CONN* filtering will be crucially needed in noisy cases. The number of representatives $m$ was set manually by examining a range of values. Clustering accuracies and the number of edges used for clustering, were tracked and averaged over 100 runs.

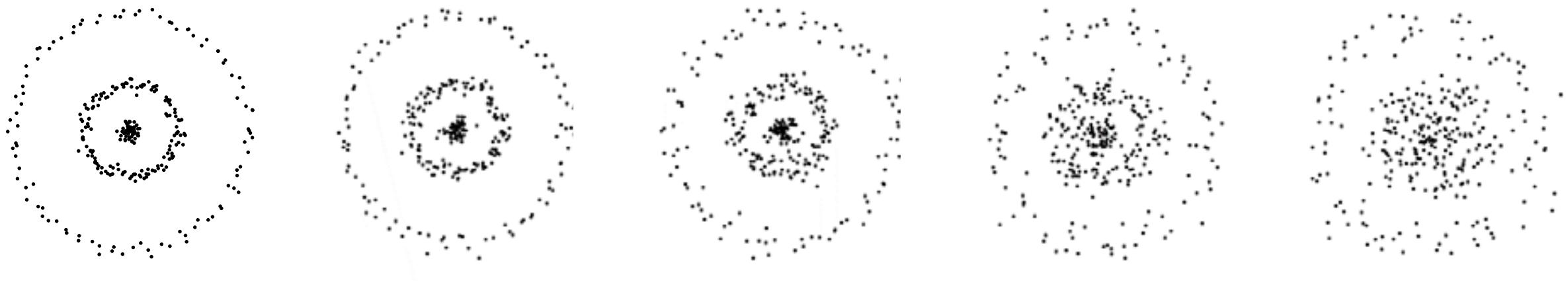

σ = 0 σ = 0.01 σ = 0.02 σ = 0.03 σ = 0.04

**Fig. 5. Different noise levels injected into synthetic dataset.**

**Table 1. Testing the original *CONN* and the proposed filtered graphs on synthetic datasets. White shaded rows are clustering accuracies, and grey shaded rows are number of edges. All are averaged over 100 runs.**

| | m | σ | CONN | $CONN_G$ | $CONN_{L1}$ | $CONN_{L2}$ |
|---|---|---|---|---|---|---|
| 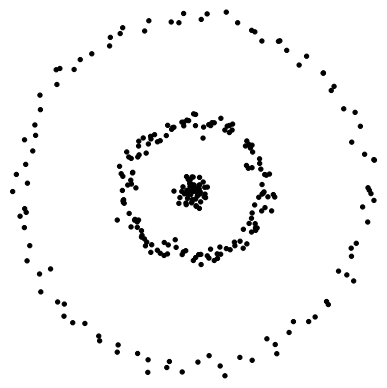 | 32 | 0.01 | **96.65 ± 10.6** | 88.68 ± 16.3 | 92.72 ± 12.3 | 89.44 ± 16.3 |
| | | | 33.48 ± 2.2 | 32.75 ± 1.2 | 31.02 ± 1.0 | 32.47 ± 1.1 |
| | | 0.02 | 87.60 ± 15.6 | 82.95 ± 16.6 | **91.66 ± 12.6** | 85.13 ± 16.3 |
| | | | 35.03 ± 2.2 | 34.05 ± 1.6 | 31.96 ± 1.1 | 33.83 ± 1.5 |
| | | 0.03 | 63.64 ± 3.5 | 65.92 ± 4.0 | **67.74 ± 7.5** | 65.09 ± 4.9 |
| | | | 42.49 ± 2.4 | 41.70 ± 2.6 | 34.00 ± 2.0 | 39.73 ± 2.7 |
| | | 0.04 | 62.67 ± 3.7 | 61.75 ± 4.7 | **63.30 ± 6.1** | 61.85 ± 5.3 |
| | | | 49.13 ± 2.5 | 47.76 ± 3.1 | 36.36 ± 2.0 | 45.20 ± 2.9 |
| 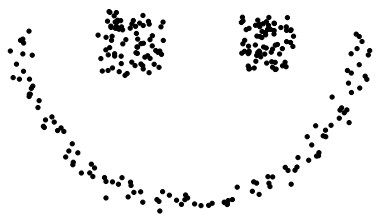 | 16 | 0.01 | **96.20 ± 6.4** | 93.50 ± 9.2 | 95.45 ± 10.6 | 96.17 ± 7.2 |
| | | | 23.71 ± 1.6 | 23.83 ± 1.7 | 21.19 ± 1.0 | 23.10 ± 1.5 |
| | | 0.02 | 91.50 ± 6.2 | 89.52 ± 8.9 | 85.56 ± 17.0 | **91.62 ± 8.7** |
| | | | 28.29 ± 1.8 | 28.03 ± 1.7 | 21.75 ± 1.3 | 26.35 ± 1.7 |
| | | 0.03 | 85.09 ± 3.7 | 84.96 ± 3.8 | **88.52 ± 11.1** | 86.91 ± 4.1 |
| | | | 32.55 ± 1.8 | 32.53 ± 1.8 | 23.41 ± 1.6 | 29.79 ± 1.8 |
| | | 0.04 | 79.29 ± 2.9 | 79.31 ± 3.4 | 80.11 ± 8.2 | **80.70 ± 4.6** |
| | | | 37.54 ± 1.9 | 37.50 ± 2.0 | 26.46 ± 1.8 | 34.22 ± 2.2 |
| 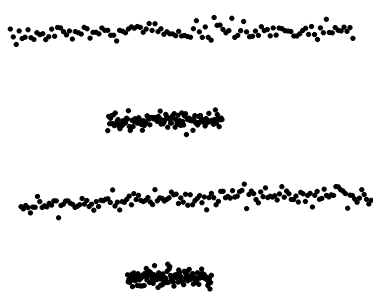 | 32 | 0.01 | **99.19 ± 5.0** | 98.47 ± 5.7 | 98.17 ± 6.3 | 98.13 ± 6.4 |
| | | | 28.18 ± 0.4 | 28.23 ± 0.7 | 27.98 ± 0.4 | 28.09 ± 0.5 |
| | | 0.02 | **98.95 ± 6.0** | 96.91 ± 9.4 | 96.69 ± 8.3 | 97.71 ± 8.3 |
| | | | 31.49 ± 1.8 | 31.62 ± 1.7 | 28.76 ± 0.9 | 30.95 ± 1.5 |
| | | 0.03 | 94.06 ± 9.1 | 92.14 ± 9.6 | 90.81 ± 13.4 | **95.94 ± 7.6** |
| | | | 39.60 ± 2.7 | 39.46 ± 2.7 | 31.42 ± 2.0 | 37.50 ± 2.5 |
| | | 0.04 | 71.85 ± 9.0 | 72.64 ± 8.4 | **80.68 ± 8.9** | 76.12 ± 9.1 |
| | | | 50.34 ± 2.2 | 50.75 ± 2.5 | 37.13 ± 2.1 | 46.88 ± 2.5 |

Table 1 shows the results of running multiple filtering schemes on synthetic data. For the first dataset containing three rings, *CONN* with no filtering was by far the best performer when the data was clean keeping a 4% advantage from its closest competitor (*$CONN_{L1}$*). This advantage was not maintained in the consequent runs when noise levels increase. *$CONN_{L1}$* scored the highest accuracies when σ was 0.02, 0.03, and 0.04. Other filtering methods (e.g., *$CONN_G$* and *$CONN_{L2}$*) could not beat *$CONN_{L1}$* because it was the most aggressive filtering scheme. Another observation for this dataset was the number of edges that grows as the noise gets higher. This is explained by the movement of points where representatives need more edges to be connected. *$CONN_{L1}$* maintained the lowest number of edges in increasing noise levels.

For the remaining datasets (smile and 4 lines), we experienced a similar observation as the first dataset. *CONN* was the best performer when the data was clean, however, it handed the first position to other methods as noise levels increase. For the smile dataset *$CONN_{L2}$* was the highest in consequent runs. *$CONN_{L1}$* tends to break edges in the middle of the arc representing the mouth because of density variations. For the 4 lines dataset, it requires σ to be 0.03 for other methods to beat *CONN*, with *$CONN_{L2}$* has 2% advantage. Nevertheless, with extreme noise (σ = 0.04), *$CONN_{L1}$* has 9% advantage over *CONN*.

### *4.2. Real images*

Four image segmentation datasets, retrieved from three sources, were used in this experiment. Berkeley segmentation dataset (BSDS500) was introduced by (Arbelaez et al., 2011). The 500 images contain different scenes to provide more diversity for the dataset. BSDS500 includes three evaluation metrics: segmentation covering (covering), rand index (RI), and variation of information (VI). For a comprehensive review on these metrics, we refer the reader to the original article by (Arbelaez et al., 2011). A rich and annotated dataset was developed at Graz University of Technology and introduced by (Opelt et al., 2004). It has 900 images for three objects: bike, car, and a person. It also includes the ground truth masks for all images. However, it does not include any evaluation metrics. Therefore, we converted all ground truth masks to be compatible with BSDS500 evaluation metrics. The last two datasets were retrieved from a repository by Weizmann Institute of Science (Alpert et al., 2007). The first dataset has a single object in the foreground, while the second has two objects. The supplied evaluation code computes F-measure for the foreground class.

An influential parameter that must be set carefully in real images is the number of representatives *m*. We carried out a small experiment showing the number of color patterns in 25000 images retrieved from MIRFLICKR-25000 (Huiskes and Lew, 2008). The histogram in Fig. 6 shows that most of color patterns would be sufficiently represented by 100 representatives.

Table 2 shows the scores of competing methods on BSDS500. $CONN_{L1}$ was the best performer in terms of segmentation covering and variation of information. This was coupled with a considerable reduction in number of edges (100 edges less than *CONN*). Higher scores and less edges nominate $CONN_{L1}$ to be the efficient method for BSDS500. The same observation continues to persist in GRAZ dataset (Table 3), where $CONN_{L1}$ got higher scores with less edges.

For Weizmann datasets (Table 4 and Table 5), $CONN_G$ got the highest scores. However, this graph was not very different from *CONN*, as one can tell from the number of edges. $CONN_{L1}$ has the lowest number of edges and its performance deviated from *CONN* by +0.2% and −3.0% for both datasets. This leaves the choice for the user to balance the tradeoff between performance (clustering accuracy) and memory footprint (number of edges).

### 4.3. UCI repository datasets

Three datasets were retrieved from UCI machine learning repository. These datasets were used in an experiment by (Tasdemir, 2012). The number of representatives was set similar to the values selected in that paper. By looking at Table 6, *CONN* produces the best performances in terms of clustering accuracy for datasets: image segmentation and pen digits. However, it falls short in Statlog dataset, handing the first position to $CONN_{L1}$.

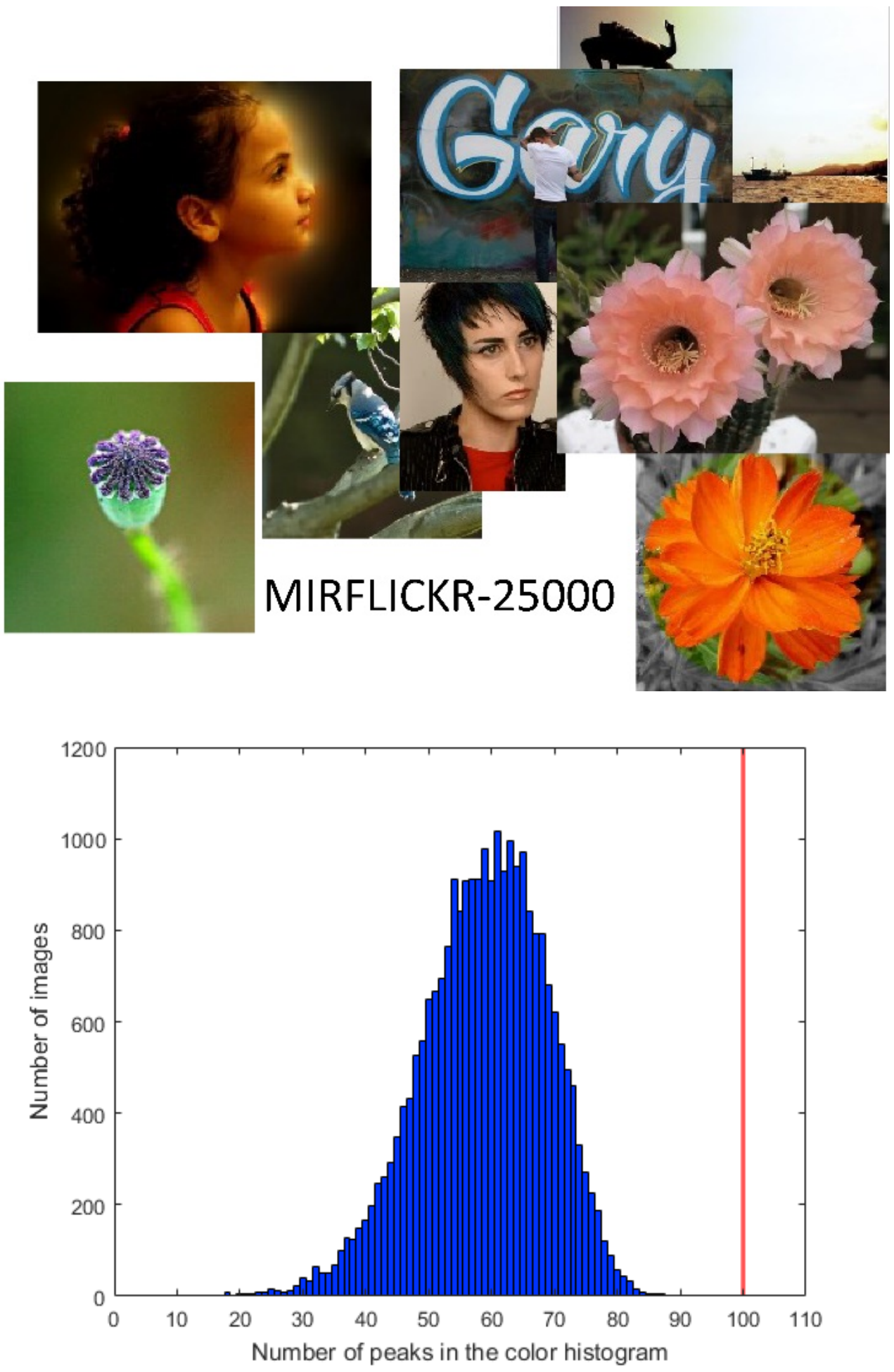

**Fig. 6. Setting the number of representatives *m* to 100 would sufficiently cover color patterns extracted from 25000 images (best viewed in color).**

**Table 2. Testing the original *CONN* and the proposed filtered graphs on BSDS500.**

| | m | CONN | $CONN_G$ | $CONN_{L1}$ | $CONN_{L2}$ |
|---|---|---|---|---|---|
| covering | 100 | 0.39 | 0.39 | **0.41** | 0.40 |
| RI | | **0.68** | 0.67 | **0.68** | **0.68** |
| VI | | 2.67 | 2.68 | **2.60** | 2.67 |
| *E* | | 280.7 ± 40.9 | 280.5 ± 41.0 | 196.3 ± 33.4 | 261.7 ± 38.5 |

**Table 3. Testing the original *CONN* and the proposed filtered graphs on GRAZ.**

| | m | CONN | $CONN_G$ | $CONN_{L1}$ | $CONN_{L2}$ |
|---|---|---|---|---|---|
| covering | 100 | 0.43 | 0.43 | **0.48** | 0.43 |
| RI | | 0.46 | 0.46 | **0.50** | 0.46 |
| VI | | 2.00 | 2.00 | **1.83** | 2.00 |
| *E* | | 292.4 ± 36.7 | 292.9 ± 36.5 | **200.5 ± 28.9** | 277.5 ± 35.0 |

**Table 4. Testing the original *CONN* and the proposed filtered graphs on Weizmann dataset (1 object).**

| | m | CONN | $CONN_G$ | $CONN_{L1}$ | $CONN_{L2}$ |
|---|---|---|---|---|---|
| F-measure | 100 | 68.6 ± 0.2 | **69.0 ± 0.2** | 68.8 ± 0.2 | 68.6 ± 0.2 |
| *E* | | 255.2 ± 39.0 | 255.0 ± 39.9 | 176.6 ± 29.9 | 238.0 ± 36.1 |

**Table 5. Testing the original *CONN* and the proposed filtered graphs on Weizmann dataset (2 objects).**

| | m | CONN | $CONN_G$ | $CONN_{L1}$ | $CONN_{L2}$ |
|---|---|---|---|---|---|
| F-measure | 100 | 51.72 ± 0.3 | **51.90 ± 0.3** | 48.69 ± 0.3 | 51.69 ± 0.2 |
| *E* | | 245.6 ± 47.2 | 246.0 ± 48.0 | 171.2 ± 36.4 | 230.0 ± 46.0 |

**Table 6. Testing the original *CONN* and the proposed filtered graphs on UCI datasets. White shaded rows are clustering accuracies, and grey shaded rows are number of edges. All are averaged over 100 runs.**

| | m | CONN | CONN$_{G}$ | CONN$_{L1}$ | CONN$_{L2}$ |
|---|---|---|---|---|---|
| Image Segmentation | 40 | **53.62 ± 4.9** | 52.58 ± 5.1 | 50.78 ± 6.0 | 51.87 ± 5.5 |
| | | 96.00 ± 6.5 | 94.42 ± 6.6 | 69.52 ± 5.1 | 87.45 ± 5.9 |
| *n = 2391* | 75 | 44.17 ± 4.3 | 44.86 ± 4.6 | 42.18 ± 6.4 | **44.93 ± 4.3** |
| *d = 19* | | 159.25 ± 7.5 | 159.59 ± 7.1 | 116.84 ± 6.5 | 145.59 ± 6.8 |
| *C = 7* | 150 | **40.71 ± 5.2** | 39.61 ± 4.3 | 37.87 ± 5.2 | 39.81 ± 4.3 |
| | | 291.21 ± 8.8 | 288.59 ± 9.0 | 217.43 ± 7.5 | 265.42 ± 8.6 |
| Statlog | 50 | 68.35 ± 4.4 | 68.36 ± 3.8 | **70.73 ± 3.6** | 68.86 ± 4.1 |
| | | 116.49 ± 3.2 | 116.84 ± 3.1 | 81.07 ± 3.5 | 109.75 ± 3.7 |
| *n = 6435* | 100 | 66.38 ± 2.9 | 66.06 ± 2.7 | **71.73 ± 3.9** | 67.49 ± 4.2 |
| *d = 4* | | 275.00 ± 7.2 | 274.01 ± 7.9 | 193.39 ± 7.4 | 255.62 ± 7.1 |
| *C = 6* | 200 | 65.61 ± 1.8 | 65.43 ± 1.9 | **71.19 ± 3.4** | 66.46 ± 3.3 |
| | | 623.84 ± 11.8 | 621.13 ± 11.9 | 450.88 ± 10.9 | 572.30 ± 11.6 |
| Pen Digits | 100 | **78.42 ± 5.1** | 78.05 ± 5.2 | 77.93 ± 5.4 | 77.96 ± 5.3 |
| | | 363.71 ± 10.1 | 363.84 ± 11.0 | 263.01 ± 9.4 | 337.41 ± 9.7 |
| *n = 10992* | 169 | **81.59 ± 4.3** | 81.16 ± 5.2 | 76.76 ± 5.8 | 79.48 ± 4.2 |
| *d = 16* | | 585.45 ± 14.6 | 584.19 ± 14.5 | 424.86 ± 11.7 | 542.39 ± 15.4 |
| *C = 10* | 324 | 81.74 ± 4.3 | **83.00 ± 4.5** | 71.46 ± 10.2 | 79.85 ± 4.6 |
| | | 1104.63 ± 17.3 | 1103.25 ± 19.5 | 807.56 ± 16.6 | 1027.75 ± 18.7 |

In general, the results produced by this experiment were below the expectations since the proposed filtering schemes were lagging behind the original *CONN*. This raised a question, why the proposed filtering works well for noisy synthetic data and real images but not for UCI datasets?. To answer this question, we examined 6 UCI datasets as well as real image datasets in terms of *DCONN Balance* measure described in subsection 3.3. The outcome of this experiment is shown in Fig. 7, in which the number of representatives *m* was fixed at 100. In that figure, all 6 UCI datasets are stacked on the left side of the plot, due to their low points to dimensions ratio. Interestingly, UCI datasets have higher *DCONN Balance* (greater than 0.2) than real image datasets. Real image datasets maintained *DCONN Balance* score lower than 0.2. Another observation is that *DCONN Balance* decreases as the points to dimensions ratio increase. Given these observations we can say that the proposed method performed better on real images because their *DCONN Balance* was not too high, at the same time it was not a completely balanced graph. On the other hand, it could not outperform *CONN* on UCI data because the graphs were extremely unbalanced. In addition, UCI data have low points to dimensions ratio. This makes removing an edge will severely impact the graph structure, compared to graphs with high points to dimensions ratio.

## 5. Conclusions

*CONN* is an efficient method to construct graphs for approximate spectral clustering (ASC). It utilizes the vector quantization step to track density information needed for weighing the graph edges. This gives *CONN* the ability to draw edges synthesizing the density of the data, and ultimately highlighting potential clusters. However, in case of noisy density, *CONN* is forced to draw edges given its definition. Also, the relationships in *CONN* are undirected, where the contribution of a vertex to the weight of an edge is not clear. We breakdown the undirected edges of *CONN* into directed edges, to highlight vertices contributions and perform global filtering accordingly. Also, we defined a local filtering to cut edges that could be drawn by noisy density that violates local statistics.

The local filtering is consuming most of the proposed method computations. Reducing these computations would improve this method. Also, testing *CONN* hybrid similarity measures would contribute to the generality of this work.

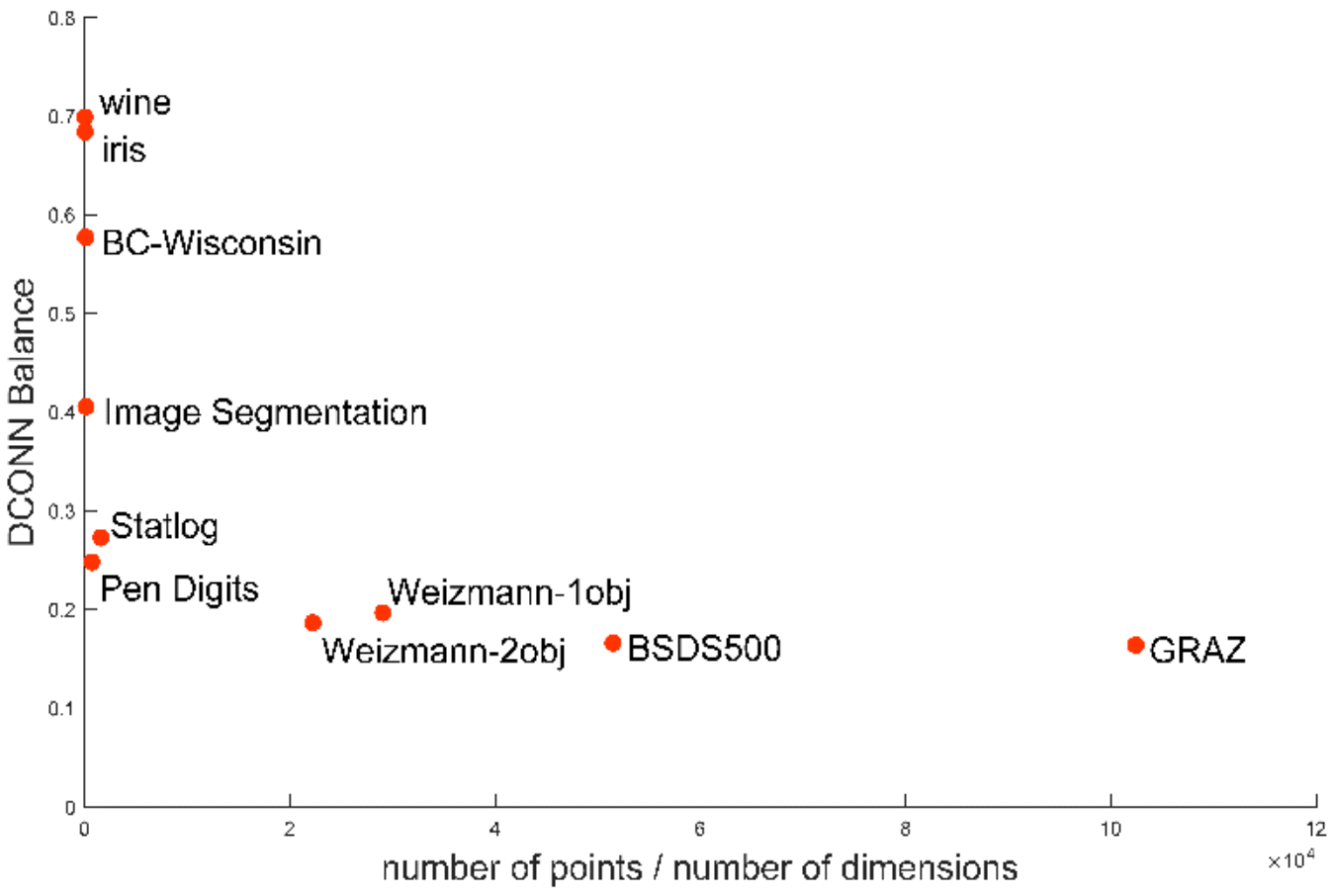

**Fig. 7. Measuring *DCONN Balance* for 6 UCI datasets and 4 real image datasets with *m* set to 100. UCI datasets have low points to dimensions ratio and high *DCONN Balance* score. Real image data have high points to dimensions ratio and moderate *DCONN Balance*.**